\documentclass{article} 
\usepackage{iclr2020_conference,times}

\usepackage{amsmath,amsfonts,bm}

\def\eqref#1{equation~\ref{#1}}

\def\1{\bm{1}}

\DeclareMathAlphabet{\mathsfit}{\encodingdefault}{\sfdefault}{m}{sl}
\SetMathAlphabet{\mathsfit}{bold}{\encodingdefault}{\sfdefault}{bx}{n}

\usepackage{hyperref}
\usepackage{url}

\usepackage{subfigure}
\usepackage{multirow}
\usepackage{booktabs}
\usepackage{multicol}
\usepackage{verbatim}

\usepackage{algorithm}
\usepackage{algorithmicx}  
\usepackage{algpseudocode}
\usepackage{bm}
\usepackage{amssymb}
\usepackage[misc]{ifsym}
\usepackage{makecell}
\usepackage{appendix}
\usepackage{graphicx}

\title{Advbox: a toolbox to generate adversarial examples that fool neural networks}

\iclrfinalcopy

\author{Dou Goodman , Hao Xin\textsuperscript{\Letter}, Wang Yang\textsuperscript{\Letter}, Xiong Junfeng \& Zhang Huan\\
Baidu X-Lab\\
Beijing, China\\
\texttt{\{wangyang62, haoxin01\}@baidu.com}
}

\begin{document}

\maketitle


\begin{abstract}
In recent years, neural networks have been extensively deployed for computer vision tasks, particularly visual classification problems, where new algorithms reported to achieve or even surpass the human performance. Recent studies have shown that they are all vulnerable to the attack of adversarial examples. Small and often imperceptible perturbations to the input images are sufficient to fool the most powerful neural networks. \emph{Advbox} is a toolbox suite to not only generate adversarial examples that fool neural networks in PaddlePaddle, PyTorch, Caffe2, MxNet, Keras, TensorFlow, but also benchmarks the robustness of machine learning models. Compared to previous work, our platform supports black box attacks on Machine-Learning-as-a-service, as well as more attack scenarios, such as Face Recognition Attack, Stealth T-shirt, and DeepFake Face Detect. 
AdvBox is openly available at https://github.com/advboxes/AdvBox. 
It now supports Python 3.
\end{abstract}

\section{Introduction}
Deep learning (DL) has made significant progress in a wide domain of machine learning (ML): image classification  \citep{krizhevsky2012imagenet,simonyan2014very,He_2016}, object detection \citep{redmon2016you,redmon2017yolo9000}, speech recognition \citep{graves2013speech,amodei2016deep}, language translation \citep{sutskever2014sequence,bahdanau2014neural}, voice synthesis \cite{oord2016wavenet,shen2018natural}. 

Szegedy et al. first generated small perturbations on the images for the image classification problem and fooled state-of- the-art deep neural networks with high probability \citep{szegedy2013intriguing}. These misclassified samples were named as \emph{Adversarial Examples}. A large number of attack algorithms have been proposed, such as FGSM \citep{goodfellow2014explaining}, BIM \citep{kurakin2016adversarial}, DeepFool \citep{Moosavi_Dezfooli_2016}, JSMA \citep{papernot2016limitations}, CW \citep{carlini2017towards}, PGD \citep{aleks2017deep}.

The scope of researchers' attacks has also gradually extended from the field of computer vision \citep{fischer2017adversarial,Xie_2017,wang2019daedalus,jia2020fooling} to the field of natural language processing \citep{ebrahimi2017hotflip,li2018textbugger,gao2018black,goodman2020fastwordbug} and speech \citep{Carlini_2018,qin2019imperceptible,Yakura_2019}.

Cloud-based services offered by Amazon\footnote{https://aws.amazon.com/cn/rekognition/}, Google\footnote{https://cloud.google.com/vision/}, Microsoft\footnote{https://azure.microsoft.com}, Clarifai\footnote{https://clarifai.com} and other public cloud companies have developed ML-as-a-service tools. Thus, users and companies can readily benefit from ML applications without having to train or host their own models\citep{Hosseini_2017}. 
Unlike common attacks against web applications, such as SQL injection and XSS, there are very special attack methods for machine learning applications, e.g., \emph{Adversarial Attack}. Obviously, neither public cloud companies nor traditional security companies pay much attention to these new attacks and defenses\citep{goodman2020attacking,goodman2019cloudbased,Li_2019,goodman2019cloud,goodman2019transferability01,goodman2019hitbtransferability,goodman2020transferability,advbox2018}.

In this paper, we will focus on adversarial example attack, defense and detection methods based on our AdvBox. Our key items covered:
\begin{itemize}
    \item The basic principles and implementation ideas.
    \item Adversarial example attack, defense and detection methods.
    \item Black box attacks on Machine-Learning-as-a-service.
    \item More attack scenarios, such as Face Recognition Attack, Stealth T-shirt, and Deepfake Face Detect.
\end{itemize}

\section{Related Work}
Currently, several attack/defense platforms have been proposed, like Cleverhans \citep{papernot2016cleverhans}, FoolBox \citep{rauber2017foolbox}, ART \citep{art2018}, DEEPSEC \citep{ling2019deepsec},  etc. For a detailed comparison, see the Table~\ref{tab:RelatedWork}.

\begin{table*}[!htbp]
	\caption{Comparison of different adversarial attack/defense platforms. "$\surd$" means "support". }
	\label{tab:RelatedWork}
	\centering
	\resizebox{0.9\textwidth}{!}{
	\begin{tabular}{c|c|c|c|c|c}
		\toprule
		 & Cleverhans & FoolBox & ART & DEEPSEC & Our  \\
		\midrule
		Tensorflow\citep{abadi2016tensorflow} & $\surd$ & $\surd$ & $\surd$ &  & $\surd$  \\
		PyTorch\citep{paszke2019pytorch} & $\surd$ & $\surd$ & $\surd$ & $\surd$ & $\surd$  \\ 
		MxNet\citep{chen2015mxnet} &  & $\surd$ & $\surd$ &  & $\surd$  \\
		PaddlePaddle\footnote{https://github.com/paddlepaddle/paddle} &  &  &  &  & $\surd$  \\ 
		Adversarial Attack & $\surd$ & $\surd$ & $\surd$ & $\surd$  & $\surd$  \\ 
		Adversarial Defense & $\surd$ & $\surd$ & $\surd$ & $\surd$  & $\surd$  \\ 
		Robustness Evaluation &  & $\surd$ & $\surd$ & $\surd$  & $\surd$  \\ 
		Adversarial Detection &  & $\surd$ & $\surd$ & $\surd$  & $\surd$  \\ 
		Attack on ML-as-a-service &  &  & &   & $\surd$  \\ 
		Actual attack scenario &  &  & &   & $\surd$  \\ 
		\bottomrule
	\end{tabular}
	}
\end{table*}

\section{Adversarial Attack}
\subsection{Problem Formulation}
The function of a pre-trained classification model $F$, e.g. an image classification or image detection model, is mapping from input set to the label set. For a clean image example $O$, it is correctly classified by $F$ to ground truth label $y \in Y$, where $Y$ including $\{1,2, \ldots, k\}$ is a label set of $k$ classes. 
An attacker aims at adding small perturbations in $O$ to generate adversarial example $ADV$, so that $F(ADV) \neq F(O)$, where $D(ADV,O)<\epsilon$, $D$ captures the semantic similarity between $ADV$ and $O$, $\epsilon$ is a threshold to limit the size of perturbations. For computer vision, $D$ usually stands for Perturbation Measurement.

\subsection{Perturbation Measurement}

$l_p$ measures the magnitude of perturbation by $p$-norm distance:
\begin{equation}
    \|x\|_{p}=\left(\sum_{i=1}^{n}\left\|x_{i}\right\|^{p}\right)^{\frac{1}{p}}
\end{equation}
$l_0$, $l_2$, $l_{\infty}$ are three commonly used $l_p$ metrics. $l_0$ counts the number of pixels changed in the adversarial examples; $l_2$ measures the Euclidean distance between the adversarial example and the original sample; $l_{\infty}$ denotes the maximum change for all pixels in adversarial examples.

\section{AdvBox}
\subsection{Overview}

\emph{Advbox} is a toolbox suite not only generate adversarial examples that fool neural networks in PaddlePaddle\footnote{https://github.com/paddlepaddle/paddle}, PyTorch\citep{paszke2019pytorch}, Caffe2\footnote{https://caffe2.ai/}, MxNet\footnote{http://mxnet.incubator.apache.org/}, Keras, TensorFlow\citep{abadi2016tensorflow} but also benchmarks the robustness of machine learning models.

\subsection{Structure}

\emph{Advbox} is based on Python\footnote{https://www.python.org/} and uses object-oriented programming.

\subsubsection{Attack Class}

Advbox implements several popular adversarial attacks which search adversarial examples. Each attack method uses a distance measure(L1, L2, etc.) to quantify the size of adversarial perturbations. Advbox is easy to craft adversarial example as some attack methods could perform internal hyperparameter tuning to find the minimum perturbation. The code is implemented as $advbox.attack$.

\subsubsection{Model Class}

Advbox implements interfaces to Tensorflow\citep{abadi2016tensorflow}, PyTorch\citep{paszke2019pytorch}, MxNet\citep{chen2015mxnet}, and PaddlePaddle\footnote{https://github.com/paddlepaddle/paddle}. Additionally, other deep learning framworks such as TensorFlow can also be defined and employed. The module is use to compute predictions and gradients for given inputs in a specific framework. 

AdvBox also supports GraphPipe\footnote{https://github.com/oracle/graphpipe}, which shields the underlying deep learning platform. Users can conduct black box attack on model files generated by Caffe2\footnote{https://caffe2.ai/}, CNTK\footnote{https://docs.microsoft.com/en-us/cognitive-toolkit/}, MATLAB\footnote{https://www.mathworks.com/products/deep-learning.html} and Chainer\footnote{https://chainer.org/} platforms. The code is implemented as $advbox.model$.

\subsubsection{Adversary Class}

Adversary contains the original object, the target and the adversarial examples. It provides the misclassification as the criterion to accept a adversarial example. The code is implemented as $advbox.adversary$.

\subsection{Adversarial Attack}
\label{sec:AdversarialAttackMitigation}
We support 6 attack algorithms which are included in adversarialbox, one of the sub-components of AdvBox. The adversarialbox is based on 
FoolBox v1~\cite{rauber2017foolbox}. 

\begin{itemize}
    \item \textbf{FGSM\citep{goodfellow2014explaining}}
    \item \textbf{BIM\citep{kurakin2016adversarial}}
    \item \textbf{DeepFool\citep{Moosavi_Dezfooli_2016}}
    \item \textbf{JSMA\citep{papernot2016limitations}}
    \item \textbf{CW\citep{carlini2017towards}}
    \item \textbf{PGD\citep{aleks2017deep}}
\end{itemize}

The code is implemented as $advbox.attack$. JSMA are often used as a baseline $l_0$ attack algorithm. CW are often used as a baseline $l_2$ attack algorithm. FGSM and PGD are often used as a baseline $l_{\infty}$ attack algorithm.

\subsection{Adversarial Attack Mitigation}
\label{sec:AdversarialAttackMitigation}
Advbox supports 6 defense algorithms:

\begin{itemize}
    \item \textbf{Feature Squeezing\citep{xu2017feature}}
    \item \textbf{Spatial Smoothing\citep{xu2017feature}}
    \item \textbf{Label Smoothing\citep{xu2017feature}}
    \item \textbf{Gaussian Augmentation\citep{zantedeschi2017efficient}}
    \item \textbf{Adversarial Training\citep{madry2017towards}}
    \item \textbf{Thermometer Encoding\citep{buckman2018thermometer}}
\end{itemize}
The code is implemented as $advbox.defense$. Adversarial Training is often used as a baseline defense algorithm.

\subsection{Robustness Evaluation Test}
\label{sec:RobustnessEvaluationTest}

We independently developed a sub-project \emph{ Perceptron}\footnote{https://github.com/advboxes/perceptron-benchmark} to evaluate the robustness of the model. Perceptron is a robustness benchmark for computer vision DNN models. It supports both image classification and object detection models as AdvBox, as well as cloud APIs. Perceptron is designed to be agnostic to the deep learning frameworks the models are built on.

Perceptron provides different attack and evaluation approaches:

\begin{itemize}
    \item \textbf{CW\citep{carlini2017towards}}
    \item \textbf{Gaussian Noise\citep{hosseini2017google}}
    \item \textbf{Uniform Noise\citep{hosseini2017google}}
    \item \textbf{Pepper Noise\citep{hosseini2017google}}
    \item \textbf{Gaussian Blurs\citep{goodman2019cloud,yuan2019stealthy}}
    \item \textbf{Brightness\citep{goodman2019cloud,yuan2019stealthy}}
    \item \textbf{Rotations\citep{engstrom2017rotation}}
    \item \textbf{Bad Weather\citep{narasimhan2000chromatic}}
\end{itemize}

\section{Attack scenario}

Compared to previous work\citep{abadi2016tensorflow,rauber2017foolbox,art2018,ling2019deepsec}, our platform supports more attack scenarios, such as Face Recognition Attack, Stealth T-shirt, and DeepFake Face Detect. 

\subsection{Scenario 1: Face Recognition Attack}

\begin{figure*}[htbp]
	\centering
	
	\subfigure[Labeled as "Bill Gates"]{
		\label{fig:FaceRecognitionAttack:a}
		\includegraphics[width=1.5in]{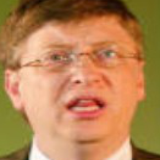}
	}%
	\subfigure[Labeled as "Michael Jordan"]{
		\label{fig:FaceRecognitionAttack:b}
		\includegraphics[width=1.5in]{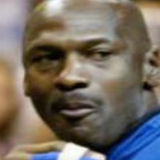}
	}%
    \subfigure[Labeled as "Michael Jordan"]{
		\label{fig:FaceRecognitionAttack:c}
		\includegraphics[width=1.5in]{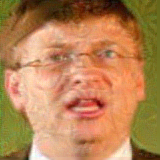}
	}%
	
	\centering
	\caption{Face Recognition Attack.}
	\label{fig:FaceRecognitionAttack}
\end{figure*}

We chose a pre-trained FaceNet\citep{schroff2015facenet} model that is a state-of-the-art and widely used face recognition system as our white box attacked model. We used gradient-based attacks methods and modify its loss function using FaceNet embedding distance.

As shown in Fig.~\ref{fig:FaceRecognitionAttack}, Fig.~\subref{fig:FaceRecognitionAttack:a} and Fig.~\subref{fig:FaceRecognitionAttack:b} can be correctly identified, but Fig.~\subref{fig:FaceRecognitionAttack:c} is incorrectly identified.

\subsection{Scenario 2: Stealth T-shirt}

\begin{figure*}[htbp]
	\centering
	
	\subfigure[]{
		\label{fig:Stealth:a}
		\includegraphics[width=2in]{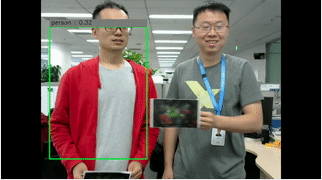}
	}%
	\subfigure[]{
		\label{fig:Stealth:b}
		\includegraphics[width=2in]{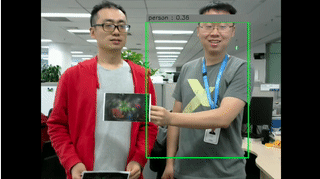}
	}%
	
	\centering
	\caption{Screenshots to demonstrate our Stealth T-shirt.}
	\label{fig:Stealth}
\end{figure*}

On Defcon China\citep{goodman2019transferability01}, we demonstrated T-shirts that can disappear under smart cameras. We open source the programs and deployment methods of smart cameras for demonstration. To raise people's awareness of techniques that can deceive deep learning models, we designed this "Stealth T-shirt" with the adversarial pattern to fool an object detector. The T-shirt is capable of hiding a person who wears it from an open-source object detector. By wearing it and showing the adversarial pattern in front of a camera and its object detector behind it, the person who wears it disappears, whereas the person who doesn't wear the T-shirt is still under object detector detection.

When the smart camera recognizes a human body in the video, it uses a green box to mark the range of the human body. Assume that the black piece of paper in the Fig.~\ref{fig:Stealth} is part of the T-shirt. As shown in Fig.~\subref{fig:Stealth:a}, the black piece of paper covered the gray man, but did not cover the man in red. The man in red was identified, and the man in gray was not identified. As shown in Fig.~\subref{fig:Stealth:b}, the black piece of paper blocked the man in red, and the man in gray was not covered. The man in gray was identified, and the man in red was not identified.

\begin{figure*}[htbp]
	\centering
	
	\subfigure[Normal]{
		\label{fig:StealthRobustness:a}
		\includegraphics[width=1.5in]{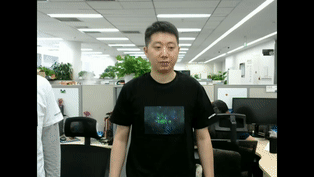}
	}%
	\subfigure[Occlusion]{
		\label{fig:StealthRobustness:b}
		\includegraphics[width=1.5in]{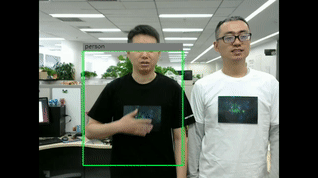}
	}%
    \subfigure[Turning]{
		\label{fig:StealthRobustness:b}
		\includegraphics[width=1.5in]{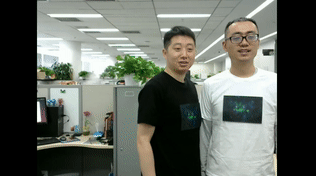}
	}%
	
	\centering
	\caption{Robustness of our Stealth T-shirt.}
	\label{fig:StealthRobustness}
\end{figure*}

As shown in the Fig.~\ref{fig:StealthRobustness}, unlike the previous work\citep{thys2019fooling,xu2019evading}, the picture we need to print in the T-shirt is smaller, facing the distortion, folding, turning, the attack effect is more robust.

\subsection{Scenario 3: DeepFake Face Detect}

We have opened the DeepFake detection capability for free, and you can remotely call our cloud detection detection api by using the Python script we provide. Deepfake is a branch of synthetic media in which a person in an existing image or video is replaced with someone else's likeness using artificial neural networks\footnote{https://en.wikipedia.org/wiki/Deepfake}. Details of our related work can refer to the conference\citep{wang2019FaceSwappingVideo,wang2019hitbDetectFakeFaces}.

\begin{figure}[htbp] 
	\centering 
	\includegraphics[width=0.6\textwidth]{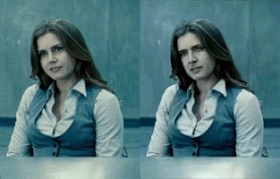}
	\caption{An example of deepfake technology: actress Amy Adams in the original (left) is modified to have the face of actor Nicolas Cage (right). } 
	\label{FL_demo} 
\end{figure}

\section*{Acknowledgement}
Thanks to every code submitter. Thanks to everyone who uses or cites \emph{AdvBox} in their papers. Especially thanks to \citet{ling2019deepsec,deng2019generate,wang2019robust} for citing us and \citet{michelini2019tour} for using us in their paper.

\bibliography{iclr2020_conference,public}
\bibliographystyle{iclr2020_conference}

\end{document}